\documentclass[sigconf]{acmart}

\usepackage{booktabs} 
\usepackage[ruled]{algorithm2e} 
\usepackage{flushend} 
\usepackage{multirow}
\usepackage{amsmath,amssymb} 
\usepackage{algorithmicx}
\usepackage{subfigure}
\usepackage{url}

\usepackage{xspace}

\makeatletter
\DeclareRobustCommand\onedot{\futurelet\@let@token\@onedot}
\def\@onedot{\ifx\@let@token.\else.\null\fi\xspace}

\makeatother

\copyrightyear{2018} 
\acmYear{2018} 
\setcopyright{acmcopyright}
\acmConference[MM '18]{2018 ACM Multimedia Conference}{October 22--26, 2018}{Seoul, Republic of Korea}
\acmPrice{15.00}
\acmDOI{10.1145/3240508.3240523}
\acmISBN{978-1-4503-5665-7/18/10}

\begin{document}
\title{Fine-Grained Representation Learning and Recognition by \\ Exploiting Hierarchical Semantic Embedding}

\author{Tianshui Chen}
\affiliation{%
  \institution{Sun Yat-sen University}
}
\email{tianshuichen@gmail.com}

\author{Wenxi Wu}
\affiliation{%
  \institution{Sun Yat-sen University}
}
\email{ngmanhei@foxmail.com}

\author{Yuefang Gao}
\affiliation{%
  \institution{South China Agricultural University}
}
\email{gaoyuefang@scau.edu.cn}

\author{Le Dong}
\affiliation{%
  \institution{University of Electronic Science and Technology of China}
}
\email{ledong@uestc.edu.cn}

\author{Xiaonan Luo}
\affiliation{%
  \institution{Guilin University of Electronic Technology}
}
\email{luoxn@guet.edu.cn}

\author{Liang Lin}
\authornote{Tianshui Chen and Wenxi Wu contribute equally to this work and share first-authorship. Corresponding author is Liang Lin.}
\affiliation{%
  \institution{Sun Yat-sen University}
}
\email{linliang@ieee.org}


\begin{abstract}
Object categories inherently form a hierarchy with different levels of concept abstraction, especially for fine-grained categories. For example, birds (Aves) can be categorized according to a four-level hierarchy of order, family, genus, and species. This hierarchy encodes rich correlations among various categories across different levels, which can effectively regularize the semantic space and thus make prediction less ambiguous. However, previous studies of fine-grained image recognition primarily focus on categories of one certain level and usually overlook this correlation information. In this work, we investigate simultaneously predicting categories of different levels in the hierarchy and integrating this structured correlation information into the deep neural network by developing a novel Hierarchical Semantic Embedding (HSE) framework. Specifically, the HSE framework sequentially predicts the category score vector of each level in the hierarchy, from highest to lowest. At each level, it incorporates the predicted score vector of the higher level as prior knowledge to learn finer-grained feature representation. During training, the predicted score vector of the higher level is also employed to regularize label prediction by using it as soft targets of corresponding sub-categories. To evaluate the proposed framework, we organize the 200 bird species of the Caltech-UCSD birds dataset with the four-level category hierarchy and construct a large-scale butterfly dataset that also covers four level categories. Extensive experiments on these two and the newly-released VegFru datasets demonstrate the superiority of our HSE framework over the baseline methods and existing competitors. 
\end{abstract}


%
%

\maketitle

\section{Introduction}
Object categories inherently form a hierarchy with different levels of concept abstraction, in which nodes closer to the root of the hierarchy refer to more abstract concepts while nodes closer to the leaves refer to finer-grained concepts. This hierarchy organization is especially important and obvious for fine-grained categories. For example, the fine-grained categories of birds (Aves) can be organized with a four-level hierarchy of order, family, genus and species, where an order consists of several families while a family consists of several genera, and so on. This category hierarchy provides very rich semantic correlations among categories across different levels, which can effectively regularize semantic space and provide extra guidance to attend more subtle regions for better recognition. For example, to recognize the fine-grained category of a given object (e.g., the species of a bird), we might first recognize its superclass (e.g., genus). Then, we prefer to concentrate on the fine-grained categories that are subject to this superclass and fixate on object parts that are more distinguishable among these fine-grained categories.

Existing methods on fine-grained image recognition (FGIR) primarily focus on classifying categories of one particular level, e.g., categorizing 200 species of birds ~\cite{lin2015bilinear,zheng2017learning} or 431 models of cars ~\cite{hu2017deep}, and usually overlook this correlation information. In this work, we simultaneously predict categories of all levels in the hierarchy, and integrate this structured correlation information into the deep neural network to progressively regularize label prediction and guide representation learning. To this, we formulate a novel Hierarchical Semantic Embedding (HSE) framework that orderly predicts the score vector of each level, from highest to lowest. At each level, it incorporates the predicted score vector of the higher level as prior knowledge to learn finer-grained feature representation. This is implemented by a semantic guided attentional mechanism that learns to fixate on more discriminative regions for better distinguishing. During training, we also utilize the predicted score vector of the higher level as soft targets to regularize the label prediction, thus that the predicted result at this level finely accords with that predicted at the higher level. 

Caltech-UCSD birds dataset ~\cite{wah2011caltech} is the most widely used benchmark for evaluating the FGIR task. To evaluate our proposed HSE framework on this benchmark, we organize the 200 bird categories with a four-level hierarchy of 13 orders, 37 families, 122 genera, and 200 species according to the ornithological systematics ~\cite{salvador2017taxonomy,remsen2016revised}. In addition, we also create a new large-scale butterfly (namely Butterfly-200) dataset that also covers four-level categories for multi-granularity image recognition. Currently, this dataset consists of 200 prevalent species of butterflies, which are grouped into 116 genera, 23 sub-families, and 5 families according to the insect taxonomy ~\cite{verovnik2013annotated,sambhu2018butterflies}. It contains 25,279 images in total and at least 30 images per species. It's worth noting that these category hierarchies can be obtained from the literature of taxonomy ~\cite{salvador2017taxonomy,verovnik2013annotated} or directly retrieved from Wikipedia conveniently, thus the methods of embedding this structured information can be easily adapted to various domains.

The major contributions of this work are concluded to three folds: 1) We formulate a novel Hierarchical Semantic Embedding (HSE) framework that integrates semantically structured information of category hierarchy into the deep neural network for FGIR. To our knowledge, this is the first work that explicitly incorporates this structured information to aid FGIR. 2) We introduce a four-level category hierarchy for the Caltech-UCSD birds dataset ~\cite{wah2011caltech} and construct a new large-scale butterfly dataset that also covers four-level categories for evaluation. To our knowledge, these two datasets are the first that involves in four-level categories in FGIR and they may benefit research on multi-granularity image recognition. 3) We conduct experiments on the two and the VegFru ~\cite{hou2017vegfru} datasets, and demonstrate the effectiveness of our proposed HSE framework over the baseline and existing state-of-the-art methods. Moreover, we also conduct ablative studies to carefully evaluate and analyze the contribution of each component of the proposed framework. \emph{The code, trained models, and dataset are available online: \url{https://github.com/HCPLab-SYSU/HSE}}.



\section{Related Work}

\subsection{Fine-grained image recognition}
Recent progress on image classification mainly benefited from the advancement of deep Convolutional Neural Networks (CNNs) ~\cite{lecun1998gradient,krizhevsky2012imagenet,simonyan2014very,he2016deep,chen2018learning,chen2016disc} that learned powerful feature representation via stacking multiple nonlinear transformations. To adapt the deep CNNs for handling the FGIR task, a bilinear model ~\cite{lin2015bilinear} was proposed to compute high-order image representation that captured local pairwise interactions between features generated by two independent sub-networks, but the bilinear feature is extremely high-dimensional, making it impractical for subsequent analysis. To reduce the feature dimension while keeping comparable performance on FGIR task, Gao et al. ~\cite{gao2016compact} developed a compact model that approximates bilinear feature with the polynomial kernels. Kong et al. ~\cite{kong2016low} proposed classifier co-decomposition to further compress the bilinear model.

To better capture subtle visual difference among sub-ordinate categories, a series works ~\cite{zhang2014part,huang2016part,zhang2016spda} were also proposed to leverage extra supervision of bounding boxes and parts to locate discriminative regions. However, the heavy involvement of manual annotations prevents these methods from application to large-scale real-world problems. Recently, visual attention models ~\cite{mnih2014recurrent,chen2018recurrent,wang2017multi,liu2018crowd} were intensively proposed to automatically search the informative regions and various works successfully applied this technique to FGIR ~\cite{liu2016fully,fu2017look,zheng2017learning,jaderberg2015spatial}. Liu et al. ~\cite{liu2016fully} formulated a reinforcement learning framework to adaptively glimpse local regions regarding discriminative object parts and trained the framework using a greedy reward strategy with image-level labels. Zheng et al. ~\cite{zheng2017learning} introduced a multi-attention convolutional neural network that learned channel grouping for parts localization, and aggregated features from the located regions as well as the global object for classification. These works learned to locate informative regions merely based on image content by the self-attention mechanism. In contrast, some works also introduced extra guidance to learn more meaningful and semantic-related regions to aid FGIR. For example, Liu et al., ~\cite{liu2017localizing,chen2018knowledge} introduce part-based attribute to guide learning more discriminative features for fine-grained bird recognition. Similarly, He et al. ~\cite{he2017fine} further utilized more detailed language descriptions to help mine discriminative parts or characteristics.

Our framework is also related to some existing works that exploit category hierarchy. For example, Srivastava et al. ~\cite{srivastava2013discriminative} exploited class hierarchy prior to transfer knowledge among similar lower-level classes for transfer learning. Jia et al. ~\cite{deng2014large} proposed a probabilistic classification model based on a hierarchy and exclusion graph to capture label relations of mutual exclusion, overlap, and subsumption for object classification. Works ~\cite{wang2016cnn,chen2018recurrent} utilized an RNN to model label co-occurrence dependencies for multi-label recognition. In contrast to these methods that merely model dependencies on label space, our HSE framework introduces the hierarchical information to progressively regularize label prediction and simultaneously guide learning finer-grained feature representation. Besides, using predicted results of the higher level as soft targets for label regularization can distill knowledge learned from the high level to lower level, which is also original compared with these methods.

\begin{figure*}[!t]
   \centering{}
   \includegraphics[width=0.90\linewidth]{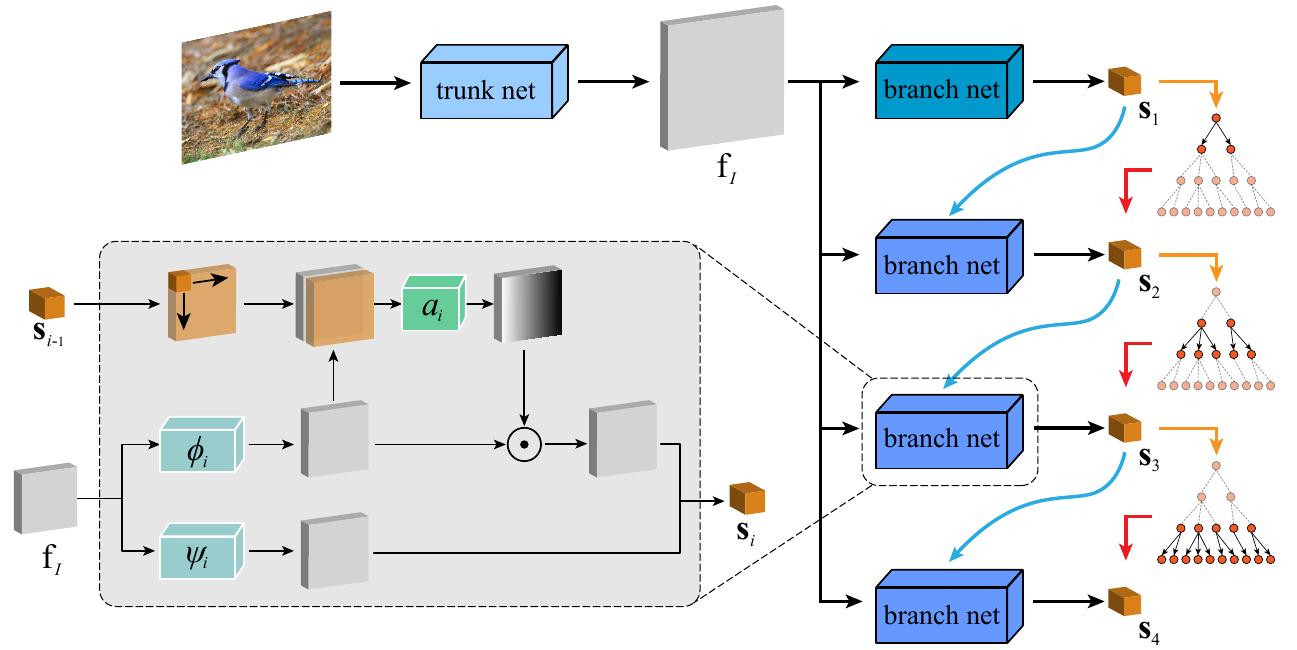}
   \caption{An overall pipeline of our proposed hierarchical semantic embedding framework. It employs a trunk network to extract image features and subsequently utilizes a branch network to predict the categories of each level. At each level, it incorporates the predicted score vector to guide learning finer-grained feature and simultaneously regularizes label prediction during training.}
   \label{fig:framework}
\end{figure*}

\subsection{Fine-grained image datasets}
In the past decade, datasets of FGIR have intensively emerged across various domains ranging from man-made objects to natural plants or animals, including FGVC-Aircraft ~\cite{maji2013fine}, Stanford Cars ~\cite{krause20133d}, Caltech-UCSD birds ~\cite{wah2011caltech}, Stanford Dogs ~\cite{khosla2011novel}, Oxford Flowers ~\cite{nilsback2008automated}, to name a few. As a representative dataset that was widely used in previous FGIR works ~\cite{liu2017localizing,gao2016compact,he2017fine}, Caltech-UCSD birds dataset contained 11,788 images and covered 200 species of birds. These datasets significantly evolved the research of FGIR, but they primarily focus on categories of one certain level, e.g., Caltech-UCSD birds with 200 species of birds and Stanford Dogs with 120 breeds of dogs. More recently, there also released some datasets that involved categories of multiple levels, like CompCars ~\cite{yang2015large}, Boxcars ~\cite{sochor2016boxcars}, Cars-333 ~\cite{xie2014hyper} with three-level car categories of make, model, and year, and VegFru ~\cite{hou2017vegfru} with 25 upper-level categories and 292 sub-ordinate classes of vegetables and fruits. These datasets mainly include man-made vehicles ~\cite{yang2015large,sochor2016boxcars,xie2014hyper} and domestic food materials ~\cite{hou2017vegfru}. To better evaluate our proposed frameworks and increase the diversity of dataset with categories of multiple levels, we further organize the 200 bird species with four-level category hierarchy and construct a new butterfly dataset that also covers four-level categories. Besides the research on FGIR with categories of multiple levels, these two datasets have potential to benefit practical applications of wildlife recognition, protection, and discovery.

\section{HSE Framework}

In this section, we describe the proposed HSE framework in detail. Given an image, the framework first utilizes a trunk network to extract image feature maps $\mathbf{f}_I \in \mathcal{R}^{W' \times H' \times C'}$, where $W'$, $H'$ and $C'$ denote the width, height and channel number of the feature maps, respectively. Then, it orderly utilizes a small branch network to predict the score vectors of all levels, from highest to lowest. At each level, the branch network incorporates the predicted score vector of higher level as prior guidance to learn finer-grained representation via a soft attention mechanism and aggregates this representation with features learned without guidance to predict the score vector of this level. During training, we further use the predicted score vector of higher level as soft targets to regularize the label prediction, such that the predicted result at this level tends to accord with that predicted at the higher level. Since there is no guidance at the first level, we merely use the representation learned without guidance to make prediction and no label regularization is involved either. Fig. \ref{fig:framework} gives an overall illustration of the HSE framework.

Before delving deep into the formulation, we first present some notations associated with our task that will be used throughout this article. Without loss of generality, we consider the FGIR task with a category hierarchy of $L$ levels. We utilize $l_1$, $l_2$, $\dots$, $l_L$ to denote each level and $\mathbf{s}_{1}$, $\mathbf{s}_{2}$, $\dots$, $\mathbf{s}_{L}$ to denote the predicted score vectors correspondingly. $n_1$, $n_2$, $\dots$, $n_L$ are used to represent the category number for each level, respectively.

\subsection{Semantic embedding representation learning}
As we orderly predict the score vector of each level, $\mathbf{s}_{i-1}$ is given when making prediction at level $l_i$. Generally, $\mathbf{s}_{i-1}$ encodes the category that the object of the given image belongs to with a high probability at level $l_{i-1}$, and make prediction at level $l_i$ may tend to distinguish the sub-ordinate categories of this category. As discussed above, some certain parts play key roles to distinguish the sub-ordinate categories of a superclass. In this work, we take full advantage of this information by incorporating $\mathbf{s}_{i-1}$ to guide learning finer-grained feature representation at level $l_i$. Naturally, this can be implemented by a soft mechanism that learns to fixate on the discriminative regions under the guidance of $\mathbf{s}_{i-1}$.

At level $l_i$, we first map the image feature maps $\mathbf{f}_I$ to higher-level features $\hat{\mathbf{f}}_i \in \mathcal{R}^{W \times H \times C}$ via

\begin{equation}
    \hat{\mathbf{f}}_i=\phi_i(\mathbf{f}_I),
\end{equation}
where $\phi_i(\cdot)$ is a transformation that is implemented by a small network. Then, at each location $(w, h)$, we introduce a shared attentional mechanism $a_i(\cdot)$ to compute the attention coefficient vector under the guidance of $\mathbf{s}_{i-1}$ by
\begin{equation}
   \hat{\mathbf{e}}_{iwh}=a_i([\hat{\mathbf{f}}_{iwh}, \varphi_i(\mathbf{s}_{{i-1}})]),
\end{equation}
where $\hat{\mathbf{e}}_{iwh}=\{\hat{e}_{iwh1}, \hat{e}_{iwh2}, \dots, \hat{e}_{iwhC}\}$ denote the importance of each neuron of feature vector $\mathbf{f}_{iwh}$. In the equation, $\varphi_i(\cdot)$ is a linear transformation that transforms $s_{i-1}$ to a semantic feature vector. To make the coefficients easily comparable across different channels, we normalize the coefficients across all the locations of each channels $c$ using a softmax function
\begin{equation}
    e_{iwhc}=\frac{\exp(\hat{e}_{iwhc})}{\sum_{w', h'}{\exp(\hat{e}_{iw'h'c})}}.
\end{equation}
In this way, we can obtain $\mathbf{e}_{iwh}=\{e_{iwh1}, e_{iwh2}, \dots, e_{iwhC}\}$ denoting the normalized weight of each neuron of feature vector $\mathbf{f}_{iwh}$. Finally, we perform weighted average across all locations of each channel to produce the final finer-grained features
\begin{equation}
    \mathbf{f}_i=\sum_{w,h}{\mathbf{e}_{iwh} \odot \hat{\mathbf{f}}_{iwh}},
\end{equation}
where $\odot$ denotes the element-wise multiplication operation. 

As the feature vector $\mathbf{f}_i$ pays much attention to the local discriminative regions that may tend to capture subtle difference for distinguishing sub-ordinate categories of a superclass. It may ignore the overall description of the object and some background information that may provide contextual cues. Thus, we further extract a feature vector directly from the image feature maps $\mathbf{f}_I$ without guidance for complementary. Similarly, we also adopt a simple transformation $\psi_i(\cdot)$ on $\mathbf{f}_I$ by
\begin{equation}
    \hat{\mathbf{f}}'_i=\psi_i(\mathbf{f}_I),
\end{equation}
where $\hat{\mathbf{f}}'_i \in \mathcal{R}^{W \times H \times C}$. Similar to ~\cite{he2016deep}, we simply perform average pooling to obtain the feature vector 
\begin{equation}
    \mathbf{f}'_i=\frac{1}{WH}\sum_{w,h}{\hat{\mathbf{f}}'_{iwh}}.
\end{equation}

The obtained feature vectors $\mathbf{f}'_i$, $\mathbf{f}_i$ and the concatenation of them $[\mathbf{f}_i, \mathbf{f}'_i]$ are fed to three classifiers to predict the score vectors independently, which are then averaged to produce the final score vector $\mathbf{s}_i$.

\noindent\textbf{Network details. }Similar to recent FGIR works ~\cite{liu2016fully,liu2017localizing}, we implement our framework based on the widely used ResNet-50 ~\cite{he2016deep}. Specifically, we implement the trunk network with the preceding 41 convolutional layers of the ResNet-50, and the transformations of $\phi_i(\cdot)$, $\psi_i(\cdot)$ with the following 9 layers of the ResNet-50. We make the trunk network be shared across different levels to better balance prediction accuracy and computational efficiency. $\varphi_i(\cdot)$ is simply implemented by a single fully connected layer that map the $c$-dim score vector to a 1,024-dimemsion features and the attention mechanism $a_i(\cdot)$ is implemented by two stacked fully connected layers, in which the first one is $c$+1,024 to 1,024 followed by the tanh non-linear function and the second one is 1,024 to $c$. As we use the identical architecture with ResNet-50, $c$ is 2,048 in this paper.

\begin{figure}[!t]
   \centering{}
   \includegraphics[width=0.8\linewidth]{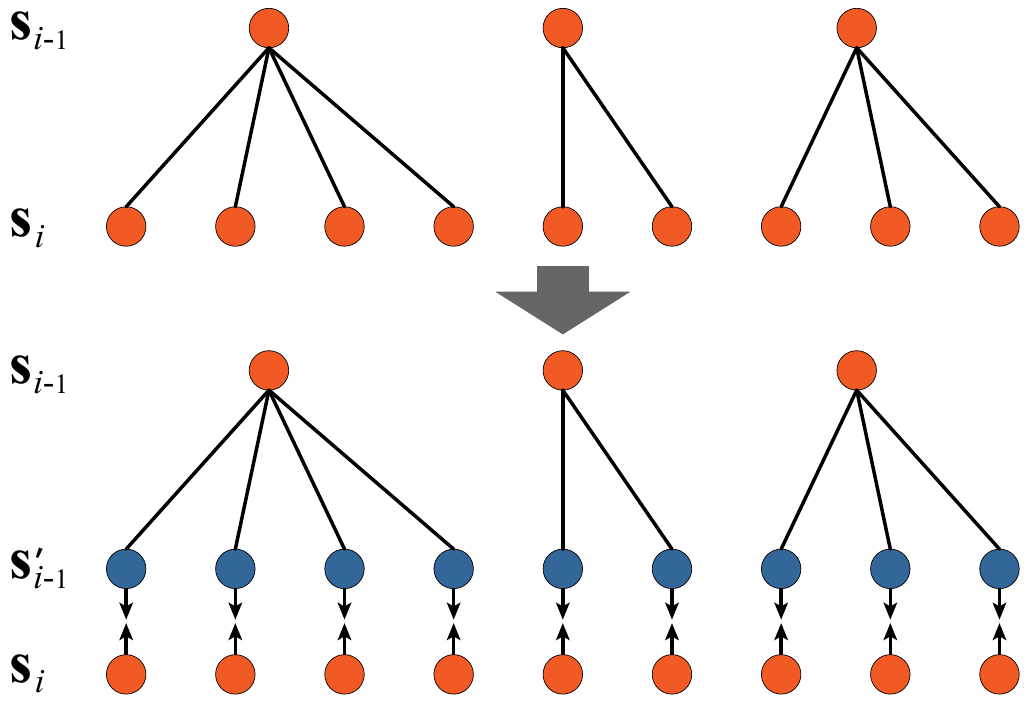} 
   \caption{An illustration of the semantic guided label regularization. Top: correlations among categories of level $l_{i-1}$ and $l_i$. Bottom: $\mathbf{s}_{i-1}$ is first extended to $\mathbf{s}'_{i-1}$ according to the structured correlations and $\mathbf{s}_i$ is pulled close to $\mathbf{s}'_{i-1}$ for regularization.}
   \label{fig:label-regularization}
\end{figure}

\subsection{Semantic guided label regularization}
The hierarchy encodes rich semantic correlations among categories across different levels. For example, the ground truth category at level $l_i$ is the child sub-category of the ground truth category at level $l_{i-1}$. This correlation information can effectively regularize semantic space and thus make prediction less ambiguous. These correlations should also be maintained among predicted categories of different levels. To this, we incorporate $\mathbf{s}_{i-1}$ as soft targets to regularize label prediction at level $l_i$.

Given the predicted score vector $\mathbf{s}_{i-1}=\{s_{i-1,1}, s_{i-1,2}, \dots, s_{i-1,n_{i-1}}\}$, a high value $s_{i-1, c}$ denotes high confidence that the object in given image belongs to category $c$ at level $l_{i-1}$, and the predicted scores for the corresponding child sub-categories at level $l_i$ should also be assigned with high values. To this, we first extend $\mathbf{s}_{i-1}$ to $\mathbf{s}'_{i-1}$ according to the structured correlations thus that $\mathbf{s}'_{i-1}$ has the same dimension as $\mathbf{s}_i$ and pull $\mathbf{s}_i$ close to $\mathbf{s}'_{i-1}$, as shown in Fig. \ref{fig:label-regularization}. Concretely, if category $c$ at level $l_{i-1}$ has $k$ child sub-categories at level $l_i$, we duplicate the score $s_{i-1, c}$ by $k$ times. Then we orderly get these duplicated scores together and re-arrange their subscripts to obtain the extended score vector $\mathbf{s}'_{i-1}=\{s'_{i-1,1}, s'_{i-1,2}, \dots, s'_{i-1,n_{i}}\}$. To make these two vectors easily comparable, we normalize them into probability distribution using the softmax function with temperature $T$
\begin{equation}
    {p}'^T_{i-1,c}=\frac{\exp(\frac{s'_{i-1,c}}{T})}{\sum_{c'}{\exp(\frac{s'_{i-1,c'}}{T})}}, 
    p^T_{i,c}=\frac{\exp(\frac{s_{i,c}}{T})}{\sum_{c'}{\exp(\frac{s_{i,c'}}{T})}},
\end{equation}
where $T$ is normally set to 1, and we use a high temperature to produce softer probability distribution over classes in our experiment. In this way, we can obtain two normalized probability distributions, i.e., $\mathbf{p}'^T_{i-1}=\{{p}'^T_{i-1,1}, {p}'^T_{i-1,2}, \dots, {p}'^T_{i-1,n_{i}}\}$ and $\mathbf{p}^T_{i}=\{p^T_{i,1}, p^T_{i,2}, \dots, p^T_{i,n_{i}}\}$, and define a regularization term as the Kullback-Leibler divergence from $\mathbf{p}^T_i$ to $\mathbf{p}'^T_{i-1}$
\begin{equation}
    \ell_i^r=D_{KL}(\mathbf{p}'^T_{i-1} || \mathbf{p}^T_i)=-\sum_{c}{p}'^T_{i-1, c}\log{\frac{p^T_{i, c}}{{p}'^T_{i-1, c}}}.
\end{equation}
As $\ell_i^r$ is defined on a single sample, we simply sum up $\ell_i^r$ over the training set to define the regularization loss term $\mathcal{L}_i^r$. As suggested in ~\cite{hinton2015distilling}, when using soft targets that have high entropy, more information can be provided than hard target per training sample, and the gradient between training samples enjoy less variance. Thus, it can be trained more steadily and using much less training samples. In our experiments, $T$ is set as 4 to produce a sufficiently soft target.

\subsection{Optimization}
Besides the regularization term, we also employ the cross-entropy loss with the correct labels as the objective function. We first normalize the predicted score vector using exactly the same logits in softmax function but at a normal temperature of 1, expressed as 
\begin{equation}
    p_{i,c}=\frac{\exp({s_{i,c}})}{\sum_{c'}{\exp({s_{i,c'}})}}.
\end{equation}
Then suppose the ground truth label at level $l_i$ is $c_i$, its loss can be defined as
\begin{equation}
    \ell_i^c=-\sum_{c}\mathbf{1}(c=c_i)\log{p_{i,c}},
\end{equation}
where $\mathbf{1}(\cdot)$ is the indication function that is assigned as 1 if the expression is true, and assigned as 0 otherwise. We have define the same loss for the score vectors predicted by the three classifier, respectively. Thus, each sample has four losses, and we sum up the four losses over the training set to define the classification loss $\mathcal{L}_i^c$.


The proposed framework consists of a trunk network and $L$ branch network, and it is trained using a weighted combination of the classification and regularization losses. The training process is empirically divided into two stages, i.e. level-wise training followed by joint fine tuning.

\noindent\textbf{Stage 1: Level-wise training. }
When training the branch network of level $l_i$, it needs the predicted score vector of level $l_{i-1}$ to define the regularization loss. Thus, we first train the branch networks in a level-wise manner, from level $l_1$ to $l_L$. As our framework is implemented based on the ResNet-50 ~\cite{he2016deep}, we initialize the parameters with those of the corresponding layers of ResNet-50 pre-trained on the ImageNet dataset ~\cite{deng2009imagenet}. Concretely, the parameters of the trunk network are initialized by those of the corresponding 41 convolutional layers and the parameters of the transformation $\phi_i(\cdot)$ and $\psi_i(\cdot)$ are initialized with those of the 9 corresponding layers. The parameters of other modules, including the attentional mechanism $a_i(\cdot)$, semantic mapper $\varphi_i(\cdot)$ and the three classifiers, are automatically initialized with the Xavier algorithm ~\cite{glorot2010understanding}. As the trunk network is shared by all branch networks, its parameters are kept fixed at this stage. We train the branch network of level $l_i$ with a weighted combination of the classification and regularization losses 
\begin{equation}
    \mathcal{L}_i=\mathcal{L}_i^c + \gamma \mathcal{L}_i^r,
\end{equation}
where $\gamma$ is a balance parameter. As discussed in ~\cite{hinton2015distilling}, the magnitudes of the gradients produced by $\mathcal{L}_i^r$ are scaled by $\frac{1}{T^2}$, thus it is important to multiply them by a scale of $T^2$. Thus, we set $\gamma$ as $T^2$, i.e., 16 in our experiments. Note that we merely use the classification loss $\mathcal{L}_1^c$ to train the branch network of level $l_1$, as there is no guidance to define the regularization loss term at this level. Similar to previous works ~\cite{liu2016fully,lin2015bilinear} on FGIR task, we resize the input images to $512 \times 512$ and perform randomly cropping with a size of $448 \times 448$ and their horizontal reflections for data augmentation. Then, we train the branch network using the stochastic gradient descent (SGD) algorithm with a batch size of 8, a momentum of 0.9 and a weight decay of 0.00005. The initial learning rate is set as 0.001, and it is divided by 10 when the error plateaus.

\noindent\textbf{Stage 2: Joint fine tuning. }After all branch networks are trained, we jointly fine tune the entire framework by combining the loss terms over all granularities
\begin{equation}
    \mathcal{L}=\mathcal{L}_1^c+\sum_{i=2}^L{\mathcal{L}_i}.
\end{equation}
We adopt the same strategies for data augmentation and hyper-parameter setting as Stage 1 except using a smaller initial learning rate 0.0001.

\begin{table*}[htp]
\centering
\begin{tabular}{c|c|c|c|c||c|c|c|c}
\hline
& \multicolumn{4}{c||}{CUB} &\multicolumn{4}{c}{Butterfly-200} \\
\hline
\centering  Methods  & $l_1$: order & $l_2$: family  & $l_3$: genus & $l_4$: species & $l_1$: family & $l_2$: sub-family  & $l_3$: genus & $l_4$: species\\
\hline
\hline
Baseline & 98.8 & 95.0 & 91.5 & 85.2 & 98.9 & 97.6 & 94.8 & 85.1 \\
Baseline+backtrack & 98.6 & 95.1 & 90.9 & 85.2 & 98.7 & 97.2 & 94.1 & 85.1 \\
\hline
\hline 
Ours w/o SERL  & 98.8 & 95.1 & 91.9 & 86.6 & 98.9 & 97.4 & 95.3 & 85.8\\
Ours w/o SGLR  & 98.8 & 95.6 & 92.2 & 86.7 & 98.9 & 97.6 & 95.1 & 85.5 \\
Ours (full) & \textbf{98.8} & \textbf{95.7} & \textbf{92.7} & \textbf{88.1} & \textbf{98.9} & \textbf{97.7} & \textbf{95.4} & \textbf{86.1} \\
\hline
\end{tabular}
\caption{Comparison of the accuracy (in \%) of all levels of our HSE framework, two baseline methods, and two variants of our framework that removes semantic embedding representation learning (Ours w/o SERL) and that removes semantic guided label regularization (Ours w/o SGLR) on the CUB and Butterfly-200 test sets, respectively.}
\label{table:result1}
\end{table*}

\section{Datasets}
We construct a new large-scale butterfly (Butterfly-200) dataset with four-level categories and organize the 200 bird species of the Caltech-UCSD Birds (CUB) dataset also with four-level categories. We evaluate our proposed framework, the baseline methods and the existing competitors on these two and the VegFru ~\cite{hou2017vegfru} datasets. In this section, we first introduce these three datasets.

\begin{figure}[!t]
   \centering
   \includegraphics[width=1.0\linewidth]{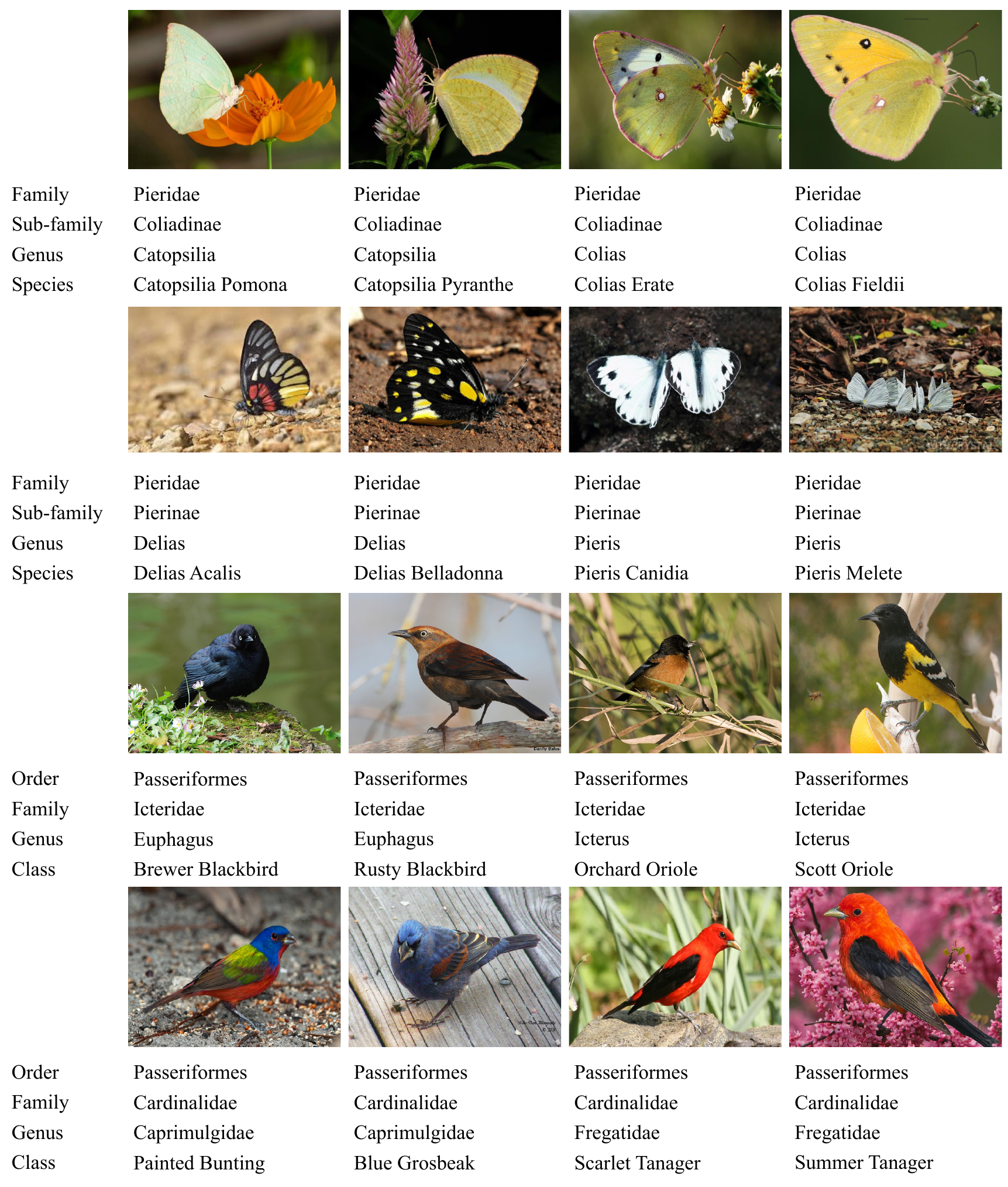} 
   \caption{Some samples and their corresponding hierarchical labels from the family of "Pieridae" in the Butterfly-200 dataset (the first two rows) and from the order of "Passeriformes" in the CUB dataset (the last two rows).}
   \label{fig:dataset}
\end{figure} 

\subsection{Butterfly-200 dataset construction}
We select 200 common species of butterflies and build the hierarchical structure with 116 genera, 23 subfamilies, and 5 families according to the insect taxonomy. The butterfly images are collected from two scenarios, natural images with the butterfly in their natural living environment and standard images with the butterfly in the form of specimens, as both are widely used in the real-world applications. The natural images are collected by searching the keywords of butterfly species names on the internet including Google, Flicker, Bing, Baidu, etc. The standard images are collected by capturing the samples in Lab. In this way, a large number of candidate images for each species are collected. To ensure the dataset highly reliable, the candidate images are carefully identified by four experts on butterflies. Currently, we have collected 25,279 butterfly images of the 200 species, with each species containing 30 images at least, which are divided into training, validation, and test set for evaluation. For each species, we randomly select 20\% for training, 20\% for validation and the rest 60\% for test, resulting in a training of 5,135 images, a validation set of 5,135 images, and a test set of 15,009 images, respectively. Figure \ref{fig:dataset} shows some samples from the family of "Pieridae" and their corresponding hierarchical labels.

\subsection{Caltech-UCSD birds dataset extenstion}
The CUB dataset ~\cite{wah2011caltech} is the most widely used benchmark for FGIR task. It covers 200 species of birds and contains 11,788 bird images that are divided into a training set of 5,994 images and a test set of 5,794 images. In this work, we build a bird taxonomy hierarchy according to the ornithological systematics, which groups the 200 species into 122 genera, 37 families, and 13 orders. We follow the standard train/test split as ~\cite{wah2011caltech} for evaluation. Figure \ref{fig:dataset} also shows some samples from the order of "Passeriformes" and their corresponding hierarchical labels.

\subsection{VegFru dataset introduction}
VegFru ~\cite{hou2017vegfru} is a newly released large-scale dataset for fine-grained vegetables and fruits recognition. It covers two-level categories of 25 upper-level categories and 292 subordinate classes. The dataset contains 160,731 images in total, including a training set of 29,200 images, a validation set of 14,600 images, and a test set of 116,931 images. Similarly, we follow this standard train/val/test splits as ~\cite{hou2017vegfru} to evaluate our HSE framework and the existing methods for fair comparison. 

\section{Experiment}

\subsection{Significance of semantic embedding}
We first implement two baseline methods that use network architecture similar to ours but do not consider the structured correlations to demonstrate the effectiveness of the proposed HSE framework. 

\noindent\textbf{Baseline. }Similar to our framework, we utilize a trunk network to extract image features and then utilize four small networks to predict the category of all levels, separately. For fair comparison, we also implement the trunk network with the preceding 41 convolutional layers of the ResNet-50 and the small network with the following 9 layers.

\noindent\textbf{Baseline+backtrack. }We utilize the baseline methods to predict the category of the finest level, and backtrack through the hierarchy to obtain the categories of the other levels. 

We compare the HSE with these two baseline methods on the CUB and Butterfly-200 datasets in Table \ref{table:result1}. Here, we present the accuracies of all levels for comprehensive comparisons. At level $l_1$, we find the HSE achieves comparable accuracies with those of the two baseline methods, as there is no semantic guidance at this level. However, at level $l_2$ to $l_4$, the HSE performs consistently better than the baseline methods on both datasets. For example on the CUB dataset, the HSE achieves accuracies of 95.7\%, 92.7\%, and 88.1\%, outperforming the baseline methods by 0.6\%, 1.2\%, and 2.9\%, respectively. It is noteworthy that the improvement is more obvious for predicting categories of finer levels, e.g., 1.2\% accuracy improvement at level $l_3$ while 2.9\% at level $l_4$ on the CUB dataset. This phenomenon suggests that incorporating semantic correction information benefits more to challenging tasks.

\begin{figure}[!t]
 \centering
 \subfigure[]{
 \includegraphics[width=0.48\linewidth]{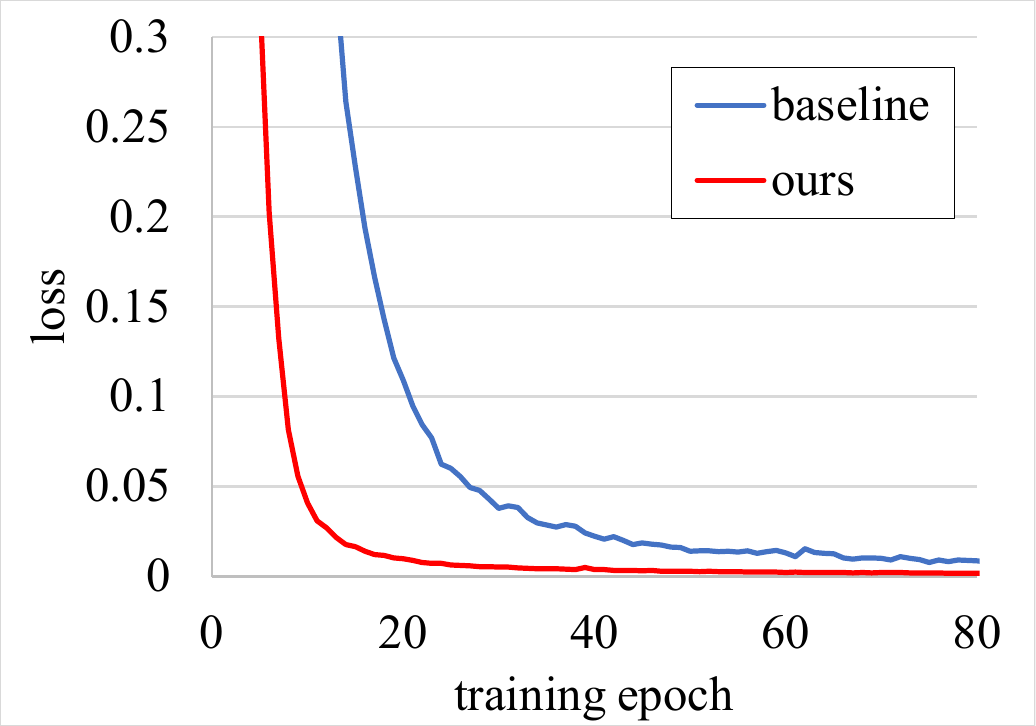}}
 \subfigure[]{
 \includegraphics[width=0.48\linewidth]{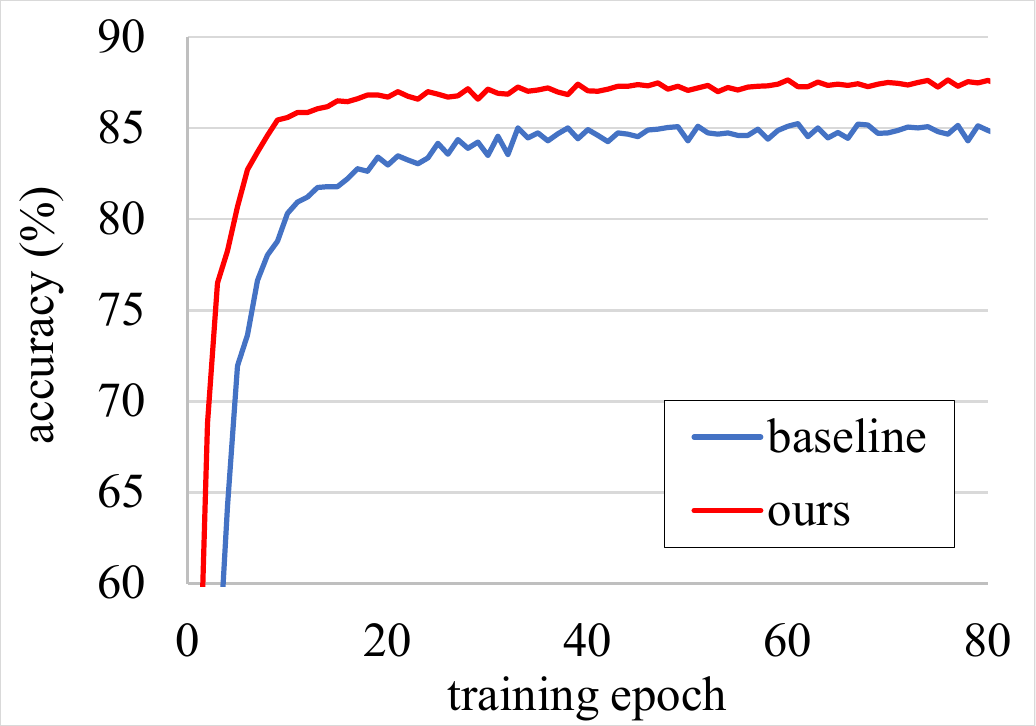}}
 \caption{Analysis of the effect of semantic embedding on network learning. These experiments are conducted on categories at level $l_4$ on the CUB dataset. (a) and (b) are the curves of loss v.s. training epoch on the training set and accuracy v.s. training epoch on the test set, respectively.}
 \label{fig:training-curve}
 \end{figure}

To delve deep into the effect of semantic embedding on network learning, we further present the curve of loss v.s. training epoch on the training set and the curve of accuracy v.s. training epoch on the test set in Fig \ref{fig:training-curve}. These experiments are conducted on recognizing the category of $l_4$ on the CUB dataset. Compared with the baseline, the HSE can be trained more stably and converged faster.

The foregoing comparisons with the baseline methods demonstrate the effectiveness of the HSE as a whole. Actually, the HSE incorporates the semantic correlation information from two aspects, i.e., semantic embedding representation learning (SERL) and semantic guided label regularization (SGLR). Here, we further conduct ablative studies to assess the actual contributions of these two components. 

\noindent\textbf{Contribution of semantic guided label regularization (SGLR). }We first evaluate the contribution of SGLR by comparing the performance with and without regularization loss. Specifically, we simply remove the regularization loss terms of each level with others keep fixed and re-train the model in an identical way. As shown in Table \ref{table:result1}, removing this term leads to an obvious drop in performance over all levels on both datasets.

\begin{figure}[!t]
\centering
\subfigure[]{
\includegraphics[width=0.48\linewidth]{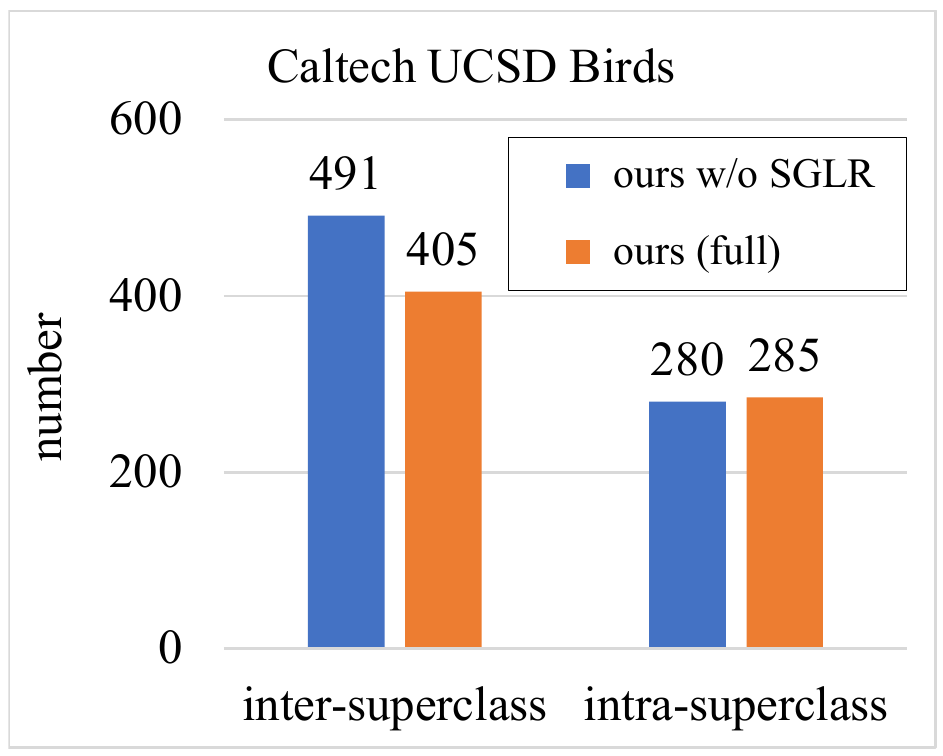}}
\subfigure[]{
\includegraphics[width=0.48\linewidth]{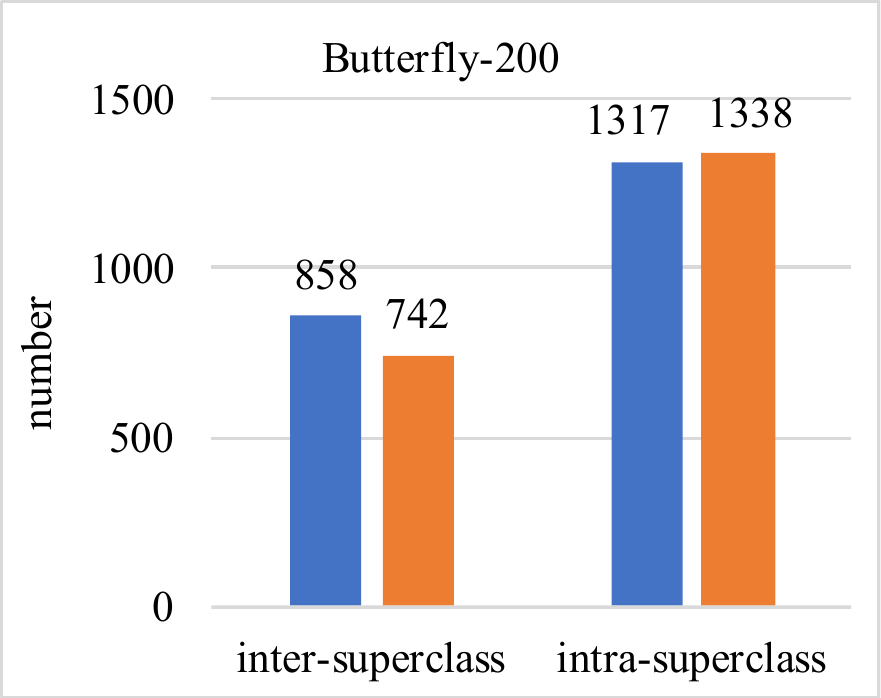}}
\caption{Sample number of inter-superclass and intra-superclass errors of our framework with and without SGLR on the (a) CUB and (b) Butterfly-200 datasets.}
\label{fig:error}
\end{figure}

We further analyze how SGLR improves the performance. When the category of an image is wrongly predicted, we denote it as an inter-superclass error if the wrongly predicted category and ground truth category do not belong to the same superclass, and denote it as an intra-superclass error if they belong to the same superclass. As discussed before, SGLR regularizes label prediction thus that the predicted category at level $l_i$ tends to be the child sub-category of the predicted category at level $l_{i-1}$. Thus, this tends to help correct the inter-superclass error. To validate this, we present the sample number of inter-superclass and the intra-superclass errors at level $l_4$ of our HSE with and without SGLR on both datasets. As shown in Fig. \ref{fig:error}, introducing SGLR mainly reduces the sample number of inter-superclass error (17.5\% relative reduction on the CUB dataset and 13.5\% on the Butterfly-200 dataset), finely in accordance with our motivation.

\noindent\textbf{Contribution of semantic embedding representation learning (SERL). }Here, we evaluate the benefit of SERL. To this, we remove the feature embedding module (i.e., $\phi_i$ and $a_i$) and simply use the feature without guidance for recognition. To ensure fair comparisons, we also re-train the model with both of the classification and regularization losses. Similarly, the performance at each level suffers from an evident drop on both datasets.

\begin{figure}[!t]
   \centering
   \includegraphics[width=0.98\linewidth]{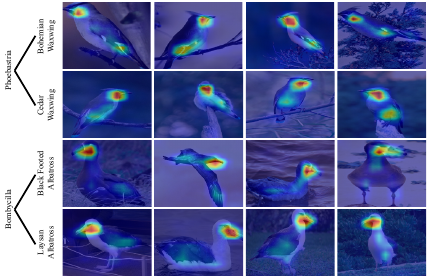} 
   \caption{Visualization of the attentional regions learned by the HSE framework. At each row, we present some samples of a specific species, and the first two species belong to the same genus while the last two belong to another genus.}
   \label{fig:vis}
\end{figure}

As discussed before, SERL helps to attend regions that help to distinguish sub-ordinate categories of the predicted superclass of the higher level. Here, we visualize the attentional regions learned by our HSE framework in Fig. \ref{fig:vis}. At each row, we present some samples of a specific species, and the first two species belong to the same genus while the last two belong to another genus. For the samples from different species of the same genus, our framework actually attends discriminative regions to better distinguish these species. For example, to differentiate the species of ``Bohemian Waxwing'' and ``Cedar Waxwing'' that belong to the genus of ``Phoebastria'', the HSE pay much attention to the throat and wing tail regions that provide most discriminative information.

\subsection{Comparison with state-of-the-art methods}
In this subsection, we compare the HSE framework with existing state-of-the-art methods on the CUB ~\cite{wah2011caltech} and VegFru ~\cite{hou2017vegfru} datasets. Here, we evaluate on recognizing the categories of the finest level (200 species on CUB and 292 subcategories on VegFru) as existing methods primarily report their results of this level.

\noindent\textbf{Comparison on Caltech-UCSD birds dataset. } CUB dataset is the most widely used benchmark for FGIR task, and most works have reported their results on this dataset. We compare our HSE framework with 17 state-of-the-art methods, including Deep Localization, Alignment and Classification (DeepLAC) ~\cite{lin2015deep}, Semantic Part Detection and Abstraction (SPDA-CNN) ~\cite{zhang2016spda}, Part-RCNN ~\cite{zhang2014part}, Part Alignment-based (PA-CNN) ~\cite{krause2015fine}, Pose Normalized CNN (PN-CNN) ~\cite{branson2014bird}, Picking Deep Filter Responses (PDFR) ~\cite{zhang2016picking}, Multiple Granularity (MG-CNN) ~\cite{wang2015multiple}, Spatial Transformer (ST-CNN) ~\cite{jaderberg2015spatial}, Bilinear-CNN (B-CNN) ~\cite{lin2015bilinear}, Compact Bilinear CNN (CB-CNN) ~\cite{gao2016compact}, Two-Level Attention Network (TLAN) ~\cite{xiao2015application}, Diverse Attention Network (DAN) ~\cite{zhao2017diversified}, Fully Convolutional Attentional Network (FCAN) ~\cite{liu2016fully}, Recurrent Attention (RA-CNN) ~\cite{fu2017look}, Combine Vision and Language (CVL) ~\cite{he2017fine}, Attribute-Guided Attention Localization (AGAL) ~\cite{liu2017localizing}, Multi-Attentional CNN (MA-CNN) ~\cite{zheng2017learning}. Among these methods, some use merely image-level labels (i.e., image-level setting), and some also use bounding box/parts annotations (i.e., box-level setting); thus we also present these information for fair and direct comparisons. 

\begin{table}[!t]
\centering
\begin{tabular}{c|c|c|c}
\hline
\centering  Methods  & BA & PA  & Acc. (\%)  \\
\hline
\hline
Part-RCNN ~\cite{zhang2014part} & $\surd$ & $\surd$ & 76.4 \\
DeepLAC ~\cite{lin2015deep} & $\surd$ & $\surd$ & 80.3 \\
SPDA-CNN ~\cite{zhang2016spda} & $\surd$ & $\surd$ & 85.1 \\
PN-CNN ~\cite{branson2014bird} & $\surd$ & $\surd$ & 85.4 \\
Part Alignment-CNN ~\cite{krause2015fine} & $\surd$ & & 82.8 \\
CB-CNN w/ bbox ~\cite{gao2016compact} & $\surd$ && 84.6 \\
FCAN w/ bbox ~\cite{liu2016fully} & $\surd$ &  & 84.7 \\
B-CNN w/ bbox ~\cite{lin2015bilinear} & $\surd$ & & 85.1 \\
AGAL w/ bbox ~\cite{liu2017localizing} & $\surd$ & & 85.5 \\
\hline
\hline
TLAN ~\cite{xiao2015application} & & & 77.9 \\
DVAN ~\cite{zhao2017diversified} & & & 79.0 \\
MG-CNN ~\cite{wang2015multiple} & & &81.7 \\
B-CNN w/o bbox ~\cite{lin2015bilinear} & & & 84.1 \\
ST-CNN ~\cite{jaderberg2015spatial} & & & 84.1 \\
FCAN w/o bbox ~\cite{liu2016fully} & &  & 84.3 \\
PDFR ~\cite{zhang2016picking} & & & 84.5 \\
CB-CNN w/o bbox ~\cite{gao2016compact} &&& 85.0 \\
RA-CNN ~\cite{fu2017look} & &  & 85.3 \\
AGAL w/o bbox ~\cite{liu2017localizing} & &  & 85.4 \\
CVL ~\cite{he2017fine} & & & 85.6 \\
MA-CNN ~\cite{zheng2017learning} & & & 86.5\\
\hline
\hline
Ours & & & \textbf{88.1} \\ 
\hline
\end{tabular}
\caption{Comparisons of our HSE framework with existing state of the arts on recognizing categories of level $l_4$ on the CUB dataset. BA and PA denote bounding box annotations and part annotations, respectively. $\surd$ indicates corresponding annotations are used during training or test. }
\label{table:cub-sota}
\end{table}


Under the box-level setting, the previous well-performing methods include PN-CNN and B-CNN that achieve accuracies of 85.4\% and 85.1\%. However, PN-CNN requires strong supervision of both human-defined bounding box and ground truth parts while B-CNN relies on a very high-dimension feature representation (250k dimensions). Under the image-level setting, most works resort to attentional model that automatically search the discriminative regions and aggregate deep features of these regions for classification. For example, MA-CNN learns to attend multiple discriminative regions, and adopt a CNN to extract the global feature from the whole and multiple part-CNNs to extract the local feature from each attentional regions. It achieves an accuracy of 86.5\%, which is the best among existing methods. Different from these methods, our HSE framework requires no bounding box and part annotations and does not use multiple CNN to extract local and global features. Instead, it embeds structure information of category hierarchy to learn fine-grained feature representation and regularize label prediction, leading to obvious performance improvement, i.e., 88.1\% in accuracy. 

Note that our HSE introduces extra guidance of the category hierarchy. However, this hierarchy can be easily obtained from the literature of taxonomy or retrieved from the Wikipedia. Besides, we also compare with existing methods that also rely on extra supervisions, like AGAL requiring attribute annotations and CVL depending on sentence description. Our HSE achieves an accuracy of 88.1\%, much better than theirs, i.e., 85.5\% and 85.6\%, respectively.

\noindent\textbf{Comparison on VegFru dataset. }VegFru is a newly released large-scale dataset for fine-grained vegetables and fruits recognition, and some works also report their results on this dataset. Here, we also present comparisons with the baseline and existing methods on this dataset in Table \ref{table:vegfru}. As shown, the HSE also significantly outperforms all these methods.

\begin{table}[htp]
\centering
\begin{tabular}{c|c}
\hline
\centering  Methods  & Acc. (\%) \\
\hline
\hline
Baseline & 87.1  \\
\hline
\hline
CB-CNN ~\cite{gao2016compact} & 82.2 \\
HybridNet ~\cite{hou2017vegfru} & 83.5 \\
\hline
\hline
Ours (full)  & \textbf{89.4} \\
\hline
\end{tabular}
\caption{Comparison of accuracy of our HSE framework, existing state-of-the-art methods, and the baseline methods on the VegFru dataset. }
\label{table:vegfru}
\end{table}

\section{Conclusion}
Fine-grained categories naturally form a hierarchy with different levels of concept abstraction, and this hierarchy encodes rich correlations among categories across different levels. In this work, we investigate simultaneously predicting categories of all levels in the hierarchy and integrating this structured correlation information into the deep neural network by developing a novel Hierarchical Semantic Embedding (HSE) framework. Specifically, the HSE orderly predicts the score vector for each level, and at each level, it incorporates the predicted score vector of the higher level to guide learning finer-grained feature representation and simultaneously regularize label prediction during training. To evaluate the HSE framework, we extend the Caltech-UCSD birds with four-level categories and construct a butterfly dataset also with four-level categories. Extensive experiments and thorough analysis on these two and the VegFru datasets demonstrate the superiority of the proposed HSE framework over the baseline methods and existing competitors.

\section*{Acknowledgement}
We would like to thank Prof. Min Wang, Associate Prof. XiaoLing Fan, Dr. Haiming Xu, and Dr. Hailing Zhuang with Department of Entomology, College of Agriculture, South China Agricultural University for their assistance in butterfly image annotations. This work was supported in part by the Chinese National Science Foundation (NSFC No. 61702196), and Science and Technology Planning Project of Guangdong Province, China (No. 2017A020208041).  This work is jointly supported by State Key Development Program under Grant 2018YFC0830103.

\bibliographystyle{ACM-Reference-Format}
\bibliography{sigproc}


\begin{thebibliography}{50}


\ifx \showCODEN    \undefined \def \showCODEN     #1{\unskip}     \fi
\ifx \showDOI      \undefined \def \showDOI       #1{#1}\fi
\ifx \showISBNx    \undefined \def \showISBNx     #1{\unskip}     \fi
\ifx \showISBNxiii \undefined \def \showISBNxiii  #1{\unskip}     \fi
\ifx \showISSN     \undefined \def \showISSN      #1{\unskip}     \fi
\ifx \showLCCN     \undefined \def \showLCCN      #1{\unskip}     \fi
\ifx \shownote     \undefined \def \shownote      #1{#1}          \fi
\ifx \showarticletitle \undefined \def \showarticletitle #1{#1}   \fi
\ifx \showURL      \undefined \def \showURL       {\relax}        \fi
\providecommand\bibfield[2]{#2}
\providecommand\bibinfo[2]{#2}
\providecommand\natexlab[1]{#1}
\providecommand\showeprint[2][]{arXiv:#2}

\bibitem[\protect\citeauthoryear{Branson, Van~Horn, Belongie, and
  Perona}{Branson et~al\mbox{.}}{2014}]%
        {branson2014bird}
\bibfield{author}{\bibinfo{person}{Steve Branson}, \bibinfo{person}{Grant
  Van~Horn}, \bibinfo{person}{Serge Belongie}, {and} \bibinfo{person}{Pietro
  Perona}.} \bibinfo{year}{2014}\natexlab{}.
\newblock \showarticletitle{Bird species categorization using pose normalized
  deep convolutional nets}.
\newblock \bibinfo{journal}{\emph{British Machine Vision Conference}}
  (\bibinfo{year}{2014}).
\newblock


\bibitem[\protect\citeauthoryear{Chen, Lin, Chen, Wu, and Luo}{Chen
  et~al\mbox{.}}{2018a}]%
        {chen2018knowledge}
\bibfield{author}{\bibinfo{person}{Tianshui Chen}, \bibinfo{person}{Liang Lin},
  \bibinfo{person}{Riquan Chen}, \bibinfo{person}{Yang Wu}, {and}
  \bibinfo{person}{Xiaonan Luo}.} \bibinfo{year}{2018}\natexlab{a}.
\newblock \showarticletitle{Knowledge-Embedded Representation Learning for
  Fine-Grained Image Recognition}. In \bibinfo{booktitle}{\emph{Proc. of
  International Joint Conference on Artificial Intelligence}}.
  \bibinfo{pages}{627--634}.
\newblock


\bibitem[\protect\citeauthoryear{Chen, Lin, Liu, Luo, and Li}{Chen
  et~al\mbox{.}}{2016}]%
        {chen2016disc}
\bibfield{author}{\bibinfo{person}{Tianshui Chen}, \bibinfo{person}{Liang Lin},
  \bibinfo{person}{Lingbo Liu}, \bibinfo{person}{Xiaonan Luo}, {and}
  \bibinfo{person}{Xuelong Li}.} \bibinfo{year}{2016}\natexlab{}.
\newblock \showarticletitle{DISC: Deep Image Saliency Computing via Progressive
  Representation Learning.}
\newblock \bibinfo{journal}{\emph{IEEE Trans. Neural Netw. Learning Syst.}}
  \bibinfo{volume}{27}, \bibinfo{number}{6} (\bibinfo{year}{2016}),
  \bibinfo{pages}{1135--1149}.
\newblock


\bibitem[\protect\citeauthoryear{Chen, Lin, Zuo, Luo, and Zhang}{Chen
  et~al\mbox{.}}{2018b}]%
        {chen2018learning}
\bibfield{author}{\bibinfo{person}{Tianshui Chen}, \bibinfo{person}{Liang Lin},
  \bibinfo{person}{Wangmeng Zuo}, \bibinfo{person}{Xiaonan Luo}, {and}
  \bibinfo{person}{Lei Zhang}.} \bibinfo{year}{2018}\natexlab{b}.
\newblock \showarticletitle{Learning a Wavelet-like Auto-Encoder to Accelerate
  Deep Neural Networks}. In \bibinfo{booktitle}{\emph{Proc. of AAAI Conference
  on Artificial Intelligence}}. \bibinfo{pages}{6722--6729}.
\newblock


\bibitem[\protect\citeauthoryear{Chen, Wang, Li, and Lin}{Chen
  et~al\mbox{.}}{2018c}]%
        {chen2018recurrent}
\bibfield{author}{\bibinfo{person}{Tianshui Chen}, \bibinfo{person}{Zhouxia
  Wang}, \bibinfo{person}{Guanbin Li}, {and} \bibinfo{person}{Liang Lin}.}
  \bibinfo{year}{2018}\natexlab{c}.
\newblock \showarticletitle{Recurrent Attentional Reinforcement Learning for
  Multi-label Image Recognition}. In \bibinfo{booktitle}{\emph{Proc. of AAAI
  Conference on Artificial Intelligence}}. \bibinfo{pages}{6730--6737}.
\newblock


\bibitem[\protect\citeauthoryear{Deng, Ding, Jia, Frome, Murphy, Bengio, Li,
  Neven, and Adam}{Deng et~al\mbox{.}}{2014}]%
        {deng2014large}
\bibfield{author}{\bibinfo{person}{Jia Deng}, \bibinfo{person}{Nan Ding},
  \bibinfo{person}{Yangqing Jia}, \bibinfo{person}{Andrea Frome},
  \bibinfo{person}{Kevin Murphy}, \bibinfo{person}{Samy Bengio},
  \bibinfo{person}{Yuan Li}, \bibinfo{person}{Hartmut Neven}, {and}
  \bibinfo{person}{Hartwig Adam}.} \bibinfo{year}{2014}\natexlab{}.
\newblock \showarticletitle{Large-scale object classification using label
  relation graphs}. In \bibinfo{booktitle}{\emph{European conference on
  computer vision}}. Springer, \bibinfo{pages}{48--64}.
\newblock


\bibitem[\protect\citeauthoryear{Deng, Dong, Socher, Li, Li, and Fei-Fei}{Deng
  et~al\mbox{.}}{2009}]%
        {deng2009imagenet}
\bibfield{author}{\bibinfo{person}{Jia Deng}, \bibinfo{person}{Wei Dong},
  \bibinfo{person}{Richard Socher}, \bibinfo{person}{Li-Jia Li},
  \bibinfo{person}{Kai Li}, {and} \bibinfo{person}{Li Fei-Fei}.}
  \bibinfo{year}{2009}\natexlab{}.
\newblock \showarticletitle{Imagenet: A large-scale hierarchical image
  database}. In \bibinfo{booktitle}{\emph{Computer Vision and Pattern
  Recognition, 2009. CVPR 2009. IEEE Conference on}}. IEEE,
  \bibinfo{pages}{248--255}.
\newblock


\bibitem[\protect\citeauthoryear{Fu, Zheng, and Mei}{Fu et~al\mbox{.}}{2017}]%
        {fu2017look}
\bibfield{author}{\bibinfo{person}{Jianlong Fu}, \bibinfo{person}{Heliang
  Zheng}, {and} \bibinfo{person}{Tao Mei}.} \bibinfo{year}{2017}\natexlab{}.
\newblock \showarticletitle{Look closer to see better: recurrent attention
  convolutional neural network for fine-grained image recognition}. In
  \bibinfo{booktitle}{\emph{IEEE Conference on Computer Vision and Pattern
  Recognition (CVPR)}}.
\newblock


\bibitem[\protect\citeauthoryear{Gao, Beijbom, Zhang, and Darrell}{Gao
  et~al\mbox{.}}{2016}]%
        {gao2016compact}
\bibfield{author}{\bibinfo{person}{Yang Gao}, \bibinfo{person}{Oscar Beijbom},
  \bibinfo{person}{Ning Zhang}, {and} \bibinfo{person}{Trevor Darrell}.}
  \bibinfo{year}{2016}\natexlab{}.
\newblock \showarticletitle{Compact bilinear pooling}. In
  \bibinfo{booktitle}{\emph{Proceedings of the IEEE Conference on Computer
  Vision and Pattern Recognition}}. \bibinfo{pages}{317--326}.
\newblock


\bibitem[\protect\citeauthoryear{Glorot and Bengio}{Glorot and Bengio}{2010}]%
        {glorot2010understanding}
\bibfield{author}{\bibinfo{person}{Xavier Glorot} {and} \bibinfo{person}{Yoshua
  Bengio}.} \bibinfo{year}{2010}\natexlab{}.
\newblock \showarticletitle{Understanding the difficulty of training deep
  feedforward neural networks}. In \bibinfo{booktitle}{\emph{Proceedings of the
  thirteenth international conference on artificial intelligence and
  statistics}}. \bibinfo{pages}{249--256}.
\newblock


\bibitem[\protect\citeauthoryear{He, Zhang, Ren, and Sun}{He
  et~al\mbox{.}}{2016}]%
        {he2016deep}
\bibfield{author}{\bibinfo{person}{Kaiming He}, \bibinfo{person}{Xiangyu
  Zhang}, \bibinfo{person}{Shaoqing Ren}, {and} \bibinfo{person}{Jian Sun}.}
  \bibinfo{year}{2016}\natexlab{}.
\newblock \showarticletitle{Deep residual learning for image recognition}. In
  \bibinfo{booktitle}{\emph{Proceedings of the IEEE conference on computer
  vision and pattern recognition}}. \bibinfo{pages}{770--778}.
\newblock


\bibitem[\protect\citeauthoryear{He and Peng}{He and Peng}{2017}]%
        {he2017fine}
\bibfield{author}{\bibinfo{person}{Xiangteng He} {and} \bibinfo{person}{Yuxin
  Peng}.} \bibinfo{year}{2017}\natexlab{}.
\newblock \showarticletitle{Fine-graind Image Classification via Combining
  Vision and Language}.
\newblock \bibinfo{journal}{\emph{Proceedings of the IEEE Conference on
  Computer Vision and Pattern Recognitions}} (\bibinfo{year}{2017}).
\newblock


\bibitem[\protect\citeauthoryear{Hinton, Vinyals, and Dean}{Hinton
  et~al\mbox{.}}{2015}]%
        {hinton2015distilling}
\bibfield{author}{\bibinfo{person}{Geoffrey Hinton}, \bibinfo{person}{Oriol
  Vinyals}, {and} \bibinfo{person}{Jeff Dean}.}
  \bibinfo{year}{2015}\natexlab{}.
\newblock \showarticletitle{Distilling the knowledge in a neural network}.
\newblock \bibinfo{journal}{\emph{arXiv preprint arXiv:1503.02531}}
  (\bibinfo{year}{2015}).
\newblock


\bibitem[\protect\citeauthoryear{Hou, Feng, and Wang}{Hou
  et~al\mbox{.}}{2017}]%
        {hou2017vegfru}
\bibfield{author}{\bibinfo{person}{Saihui Hou}, \bibinfo{person}{Yushan Feng},
  {and} \bibinfo{person}{Zilei Wang}.} \bibinfo{year}{2017}\natexlab{}.
\newblock \showarticletitle{VegFru: A Domain-Specific Dataset for Fine-grained
  Visual Categorization}. In \bibinfo{booktitle}{\emph{2017 IEEE International
  Conference on Computer Vision (ICCV)}}. IEEE, \bibinfo{pages}{541--549}.
\newblock


\bibitem[\protect\citeauthoryear{Hu, Wang, Li, and Shen}{Hu
  et~al\mbox{.}}{2017}]%
        {hu2017deep}
\bibfield{author}{\bibinfo{person}{Qichang Hu}, \bibinfo{person}{Huibing Wang},
  \bibinfo{person}{Teng Li}, {and} \bibinfo{person}{Chunhua Shen}.}
  \bibinfo{year}{2017}\natexlab{}.
\newblock \showarticletitle{Deep CNNs With Spatially Weighted Pooling for
  Fine-Grained Car Recognition}.
\newblock \bibinfo{journal}{\emph{IEEE Transactions on Intelligent
  Transportation Systems}} \bibinfo{volume}{18}, \bibinfo{number}{11}
  (\bibinfo{year}{2017}), \bibinfo{pages}{3147--3156}.
\newblock


\bibitem[\protect\citeauthoryear{Huang, Xu, Tao, and Zhang}{Huang
  et~al\mbox{.}}{2016}]%
        {huang2016part}
\bibfield{author}{\bibinfo{person}{Shaoli Huang}, \bibinfo{person}{Zhe Xu},
  \bibinfo{person}{Dacheng Tao}, {and} \bibinfo{person}{Ya Zhang}.}
  \bibinfo{year}{2016}\natexlab{}.
\newblock \showarticletitle{Part-stacked CNN for fine-grained visual
  categorization}. In \bibinfo{booktitle}{\emph{Proceedings of the IEEE
  Conference on Computer Vision and Pattern Recognition}}.
  \bibinfo{pages}{1173--1182}.
\newblock


\bibitem[\protect\citeauthoryear{Jaderberg, Simonyan, Zisserman,
  et~al\mbox{.}}{Jaderberg et~al\mbox{.}}{2015}]%
        {jaderberg2015spatial}
\bibfield{author}{\bibinfo{person}{Max Jaderberg}, \bibinfo{person}{Karen
  Simonyan}, \bibinfo{person}{Andrew Zisserman}, {et~al\mbox{.}}}
  \bibinfo{year}{2015}\natexlab{}.
\newblock \showarticletitle{Spatial transformer networks}. In
  \bibinfo{booktitle}{\emph{Advances in Neural Information Processing
  Systems}}. \bibinfo{pages}{2017--2025}.
\newblock


\bibitem[\protect\citeauthoryear{Khosla, Jayadevaprakash, Yao, and Li}{Khosla
  et~al\mbox{.}}{2011}]%
        {khosla2011novel}
\bibfield{author}{\bibinfo{person}{Aditya Khosla}, \bibinfo{person}{Nityananda
  Jayadevaprakash}, \bibinfo{person}{Bangpeng Yao}, {and}
  \bibinfo{person}{Fei-Fei Li}.} \bibinfo{year}{2011}\natexlab{}.
\newblock \showarticletitle{Novel dataset for fine-grained image
  categorization: Stanford dogs}. In \bibinfo{booktitle}{\emph{Proc. CVPR
  Workshop on Fine-Grained Visual Categorization (FGVC)}},
  Vol.~\bibinfo{volume}{2}. \bibinfo{pages}{1}.
\newblock


\bibitem[\protect\citeauthoryear{Kong and Fowlkes}{Kong and Fowlkes}{2016}]%
        {kong2016low}
\bibfield{author}{\bibinfo{person}{Shu Kong} {and} \bibinfo{person}{Charless
  Fowlkes}.} \bibinfo{year}{2016}\natexlab{}.
\newblock \showarticletitle{Low-rank Bilinear Pooling for Fine-Grained
  Classification}.
\newblock \bibinfo{journal}{\emph{arXiv preprint arXiv:1611.05109}}
  (\bibinfo{year}{2016}).
\newblock


\bibitem[\protect\citeauthoryear{Krause, Jin, Yang, and Fei-Fei}{Krause
  et~al\mbox{.}}{2015}]%
        {krause2015fine}
\bibfield{author}{\bibinfo{person}{Jonathan Krause}, \bibinfo{person}{Hailin
  Jin}, \bibinfo{person}{Jianchao Yang}, {and} \bibinfo{person}{Li Fei-Fei}.}
  \bibinfo{year}{2015}\natexlab{}.
\newblock \showarticletitle{Fine-grained recognition without part annotations}.
  In \bibinfo{booktitle}{\emph{Proceedings of the IEEE Conference on Computer
  Vision and Pattern Recognition}}. \bibinfo{pages}{5546--5555}.
\newblock


\bibitem[\protect\citeauthoryear{Krause, Stark, Deng, and Fei-Fei}{Krause
  et~al\mbox{.}}{2013}]%
        {krause20133d}
\bibfield{author}{\bibinfo{person}{Jonathan Krause}, \bibinfo{person}{Michael
  Stark}, \bibinfo{person}{Jia Deng}, {and} \bibinfo{person}{Li Fei-Fei}.}
  \bibinfo{year}{2013}\natexlab{}.
\newblock \showarticletitle{3d object representations for fine-grained
  categorization}. In \bibinfo{booktitle}{\emph{Computer Vision Workshops
  (ICCVW), 2013 IEEE International Conference on}}. IEEE,
  \bibinfo{pages}{554--561}.
\newblock


\bibitem[\protect\citeauthoryear{Krizhevsky, Sutskever, and Hinton}{Krizhevsky
  et~al\mbox{.}}{2012}]%
        {krizhevsky2012imagenet}
\bibfield{author}{\bibinfo{person}{Alex Krizhevsky}, \bibinfo{person}{Ilya
  Sutskever}, {and} \bibinfo{person}{Geoffrey~E Hinton}.}
  \bibinfo{year}{2012}\natexlab{}.
\newblock \showarticletitle{Imagenet classification with deep convolutional
  neural networks}. In \bibinfo{booktitle}{\emph{Advances in neural information
  processing systems}}. \bibinfo{pages}{1097--1105}.
\newblock


\bibitem[\protect\citeauthoryear{LeCun, Bottou, Bengio, and Haffner}{LeCun
  et~al\mbox{.}}{1998}]%
        {lecun1998gradient}
\bibfield{author}{\bibinfo{person}{Yann LeCun}, \bibinfo{person}{L{\'e}on
  Bottou}, \bibinfo{person}{Yoshua Bengio}, {and} \bibinfo{person}{Patrick
  Haffner}.} \bibinfo{year}{1998}\natexlab{}.
\newblock \showarticletitle{Gradient-based learning applied to document
  recognition}.
\newblock \bibinfo{journal}{\emph{Proc. IEEE}} \bibinfo{volume}{86},
  \bibinfo{number}{11} (\bibinfo{year}{1998}), \bibinfo{pages}{2278--2324}.
\newblock


\bibitem[\protect\citeauthoryear{Lin, Shen, Lu, and Jia}{Lin
  et~al\mbox{.}}{2015b}]%
        {lin2015deep}
\bibfield{author}{\bibinfo{person}{Di Lin}, \bibinfo{person}{Xiaoyong Shen},
  \bibinfo{person}{Cewu Lu}, {and} \bibinfo{person}{Jiaya Jia}.}
  \bibinfo{year}{2015}\natexlab{b}.
\newblock \showarticletitle{Deep lac: Deep localization, alignment and
  classification for fine-grained recognition}. In
  \bibinfo{booktitle}{\emph{Proceedings of the IEEE Conference on Computer
  Vision and Pattern Recognition}}. \bibinfo{pages}{1666--1674}.
\newblock


\bibitem[\protect\citeauthoryear{Lin, RoyChowdhury, and Maji}{Lin
  et~al\mbox{.}}{2015a}]%
        {lin2015bilinear}
\bibfield{author}{\bibinfo{person}{Tsung-Yu Lin}, \bibinfo{person}{Aruni
  RoyChowdhury}, {and} \bibinfo{person}{Subhransu Maji}.}
  \bibinfo{year}{2015}\natexlab{a}.
\newblock \showarticletitle{Bilinear cnn models for fine-grained visual
  recognition}. In \bibinfo{booktitle}{\emph{Proceedings of the IEEE
  International Conference on Computer Vision}}. \bibinfo{pages}{1449--1457}.
\newblock


\bibitem[\protect\citeauthoryear{Liu, Wang, Li, Ouyang, and Lin}{Liu
  et~al\mbox{.}}{2018}]%
        {liu2018crowd}
\bibfield{author}{\bibinfo{person}{Lingbo Liu}, \bibinfo{person}{Hongjun Wang},
  \bibinfo{person}{Guanbin Li}, \bibinfo{person}{Wanli Ouyang}, {and}
  \bibinfo{person}{Liang Lin}.} \bibinfo{year}{2018}\natexlab{}.
\newblock \showarticletitle{Crowd Counting using Deep Recurrent Spatial-Aware
  Network}. In \bibinfo{booktitle}{\emph{Proc. of International Joint
  Conference on Artificial Intelligence}}.
\newblock


\bibitem[\protect\citeauthoryear{Liu, Wang, Wen, Ding, and Lin}{Liu
  et~al\mbox{.}}{2017}]%
        {liu2017localizing}
\bibfield{author}{\bibinfo{person}{Xiao Liu}, \bibinfo{person}{Jiang Wang},
  \bibinfo{person}{Shilei Wen}, \bibinfo{person}{Errui Ding}, {and}
  \bibinfo{person}{Yuanqing Lin}.} \bibinfo{year}{2017}\natexlab{}.
\newblock \showarticletitle{Localizing by Describing: Attribute-Guided
  Attention Localization for Fine-Grained Recognition.}. In
  \bibinfo{booktitle}{\emph{AAAI}}. \bibinfo{pages}{4190--4196}.
\newblock


\bibitem[\protect\citeauthoryear{Liu, Xia, Wang, and Lin}{Liu
  et~al\mbox{.}}{2016}]%
        {liu2016fully}
\bibfield{author}{\bibinfo{person}{Xiao Liu}, \bibinfo{person}{Tian Xia},
  \bibinfo{person}{Jiang Wang}, {and} \bibinfo{person}{Yuanqing Lin}.}
  \bibinfo{year}{2016}\natexlab{}.
\newblock \showarticletitle{Fully convolutional attention localization
  networks: Efficient attention localization for fine-grained recognition}.
\newblock \bibinfo{journal}{\emph{arXiv preprint arXiv:1603.06765}}
  (\bibinfo{year}{2016}).
\newblock


\bibitem[\protect\citeauthoryear{Maji, Rahtu, Kannala, Blaschko, and
  Vedaldi}{Maji et~al\mbox{.}}{2013}]%
        {maji2013fine}
\bibfield{author}{\bibinfo{person}{Subhransu Maji}, \bibinfo{person}{Esa
  Rahtu}, \bibinfo{person}{Juho Kannala}, \bibinfo{person}{Matthew Blaschko},
  {and} \bibinfo{person}{Andrea Vedaldi}.} \bibinfo{year}{2013}\natexlab{}.
\newblock \showarticletitle{Fine-grained visual classification of aircraft}.
\newblock \bibinfo{journal}{\emph{arXiv preprint arXiv:1306.5151}}
  (\bibinfo{year}{2013}).
\newblock


\bibitem[\protect\citeauthoryear{Mnih, Heess, Graves, et~al\mbox{.}}{Mnih
  et~al\mbox{.}}{2014}]%
        {mnih2014recurrent}
\bibfield{author}{\bibinfo{person}{Volodymyr Mnih}, \bibinfo{person}{Nicolas
  Heess}, \bibinfo{person}{Alex Graves}, {et~al\mbox{.}}}
  \bibinfo{year}{2014}\natexlab{}.
\newblock \showarticletitle{Recurrent models of visual attention}. In
  \bibinfo{booktitle}{\emph{Advances in neural information processing
  systems}}. \bibinfo{pages}{2204--2212}.
\newblock


\bibitem[\protect\citeauthoryear{Nilsback and Zisserman}{Nilsback and
  Zisserman}{2008}]%
        {nilsback2008automated}
\bibfield{author}{\bibinfo{person}{Maria-Elena Nilsback} {and}
  \bibinfo{person}{Andrew Zisserman}.} \bibinfo{year}{2008}\natexlab{}.
\newblock \showarticletitle{Automated flower classification over a large number
  of classes}. In \bibinfo{booktitle}{\emph{Computer Vision, Graphics \& Image
  Processing, 2008. ICVGIP'08. Sixth Indian Conference on}}. IEEE,
  \bibinfo{pages}{722--729}.
\newblock


\bibitem[\protect\citeauthoryear{Remsen~Jr, Powell, Schodde, Barker, and
  Lanyon}{Remsen~Jr et~al\mbox{.}}{2016}]%
        {remsen2016revised}
\bibfield{author}{\bibinfo{person}{JV Remsen~Jr}, \bibinfo{person}{Alexis~FLA
  Powell}, \bibinfo{person}{Richard Schodde}, \bibinfo{person}{F~Keith Barker},
  {and} \bibinfo{person}{Scott~M Lanyon}.} \bibinfo{year}{2016}\natexlab{}.
\newblock \showarticletitle{A revised classification of the Icteridae (Aves)
  based on DNA sequence data}.
\newblock \bibinfo{journal}{\emph{Zootaxa}} \bibinfo{volume}{4093},
  \bibinfo{number}{2} (\bibinfo{year}{2016}), \bibinfo{pages}{285--292}.
\newblock


\bibitem[\protect\citeauthoryear{Salvador, Van~der Jeugd, and
  Tomotani}{Salvador et~al\mbox{.}}{2017}]%
        {salvador2017taxonomy}
\bibfield{author}{\bibinfo{person}{Rodrigo~B Salvador}, \bibinfo{person}{Henk
  Van~der Jeugd}, {and} \bibinfo{person}{Barbara~M Tomotani}.}
  \bibinfo{year}{2017}\natexlab{}.
\newblock \showarticletitle{Taxonomy of the European Pied Flycatcher Ficedula
  hypoleuca (Aves: Muscicapidae)}.
\newblock \bibinfo{journal}{\emph{Zootaxa}} \bibinfo{volume}{4291},
  \bibinfo{number}{1} (\bibinfo{year}{2017}), \bibinfo{pages}{171--182}.
\newblock


\bibitem[\protect\citeauthoryear{SAMBHU and NANKISHORE}{SAMBHU and
  NANKISHORE}{2018}]%
        {sambhu2018butterflies}
\bibfield{author}{\bibinfo{person}{HEMCHANDRANAUTH SAMBHU} {and}
  \bibinfo{person}{ALLIEA NANKISHORE}.} \bibinfo{year}{2018}\natexlab{}.
\newblock \showarticletitle{Butterflies (Lepidoptera) of Guyana: A compilation
  of records}.
\newblock \bibinfo{journal}{\emph{Zootaxa}} \bibinfo{volume}{4371},
  \bibinfo{number}{1} (\bibinfo{year}{2018}), \bibinfo{pages}{1--187}.
\newblock


\bibitem[\protect\citeauthoryear{Simonyan and Zisserman}{Simonyan and
  Zisserman}{2014}]%
        {simonyan2014very}
\bibfield{author}{\bibinfo{person}{Karen Simonyan} {and}
  \bibinfo{person}{Andrew Zisserman}.} \bibinfo{year}{2014}\natexlab{}.
\newblock \showarticletitle{Very deep convolutional networks for large-scale
  image recognition}.
\newblock \bibinfo{journal}{\emph{arXiv preprint arXiv:1409.1556}}
  (\bibinfo{year}{2014}).
\newblock


\bibitem[\protect\citeauthoryear{Sochor, Herout, and Havel}{Sochor
  et~al\mbox{.}}{2016}]%
        {sochor2016boxcars}
\bibfield{author}{\bibinfo{person}{Jakub Sochor}, \bibinfo{person}{Adam
  Herout}, {and} \bibinfo{person}{Jiri Havel}.}
  \bibinfo{year}{2016}\natexlab{}.
\newblock \showarticletitle{Boxcars: 3d boxes as cnn input for improved
  fine-grained vehicle recognition}. In \bibinfo{booktitle}{\emph{Proceedings
  of the IEEE Conference on Computer Vision and Pattern Recognition}}.
  \bibinfo{pages}{3006--3015}.
\newblock


\bibitem[\protect\citeauthoryear{Srivastava and Salakhutdinov}{Srivastava and
  Salakhutdinov}{2013}]%
        {srivastava2013discriminative}
\bibfield{author}{\bibinfo{person}{Nitish Srivastava} {and}
  \bibinfo{person}{Ruslan~R Salakhutdinov}.} \bibinfo{year}{2013}\natexlab{}.
\newblock \showarticletitle{Discriminative transfer learning with tree-based
  priors}. In \bibinfo{booktitle}{\emph{Advances in Neural Information
  Processing Systems}}. \bibinfo{pages}{2094--2102}.
\newblock


\bibitem[\protect\citeauthoryear{Verovnik and Popovi{\'c}}{Verovnik and
  Popovi{\'c}}{2013}]%
        {verovnik2013annotated}
\bibfield{author}{\bibinfo{person}{Rudi Verovnik} {and}
  \bibinfo{person}{Milo{\v{s}} Popovi{\'c}}.} \bibinfo{year}{2013}\natexlab{}.
\newblock \showarticletitle{Annotated checklist of Albanian butterflies
  (Lepidoptera, Papilionoidea and Hesperioidea)}.
\newblock \bibinfo{journal}{\emph{ZooKeys}} \bibinfo{number}{323}
  (\bibinfo{year}{2013}), \bibinfo{pages}{75}.
\newblock


\bibitem[\protect\citeauthoryear{Wah, Branson, Welinder, Perona, and
  Belongie}{Wah et~al\mbox{.}}{2011}]%
        {wah2011caltech}
\bibfield{author}{\bibinfo{person}{Catherine Wah}, \bibinfo{person}{Steve
  Branson}, \bibinfo{person}{Peter Welinder}, \bibinfo{person}{Pietro Perona},
  {and} \bibinfo{person}{Serge Belongie}.} \bibinfo{year}{2011}\natexlab{}.
\newblock \showarticletitle{The caltech-ucsd birds-200-2011 dataset}.
\newblock  (\bibinfo{year}{2011}).
\newblock


\bibitem[\protect\citeauthoryear{Wang, Shen, Shao, Zhang, Xue, and Zhang}{Wang
  et~al\mbox{.}}{2015}]%
        {wang2015multiple}
\bibfield{author}{\bibinfo{person}{Dequan Wang}, \bibinfo{person}{Zhiqiang
  Shen}, \bibinfo{person}{Jie Shao}, \bibinfo{person}{Wei Zhang},
  \bibinfo{person}{Xiangyang Xue}, {and} \bibinfo{person}{Zheng Zhang}.}
  \bibinfo{year}{2015}\natexlab{}.
\newblock \showarticletitle{Multiple granularity descriptors for fine-grained
  categorization}. In \bibinfo{booktitle}{\emph{Proceedings of the IEEE
  International Conference on Computer Vision}}. \bibinfo{pages}{2399--2406}.
\newblock


\bibitem[\protect\citeauthoryear{Wang, Yang, Mao, Huang, Huang, and Xu}{Wang
  et~al\mbox{.}}{2016}]%
        {wang2016cnn}
\bibfield{author}{\bibinfo{person}{Jiang Wang}, \bibinfo{person}{Yi Yang},
  \bibinfo{person}{Junhua Mao}, \bibinfo{person}{Zhiheng Huang},
  \bibinfo{person}{Chang Huang}, {and} \bibinfo{person}{Wei Xu}.}
  \bibinfo{year}{2016}\natexlab{}.
\newblock \showarticletitle{Cnn-rnn: A unified framework for multi-label image
  classification}. In \bibinfo{booktitle}{\emph{Computer Vision and Pattern
  Recognition (CVPR), 2016 IEEE Conference on}}. IEEE,
  \bibinfo{pages}{2285--2294}.
\newblock


\bibitem[\protect\citeauthoryear{Wang, Chen, Li, Xu, and Lin}{Wang
  et~al\mbox{.}}{2017}]%
        {wang2017multi}
\bibfield{author}{\bibinfo{person}{Zhouxia Wang}, \bibinfo{person}{Tianshui
  Chen}, \bibinfo{person}{Guanbin Li}, \bibinfo{person}{Ruijia Xu}, {and}
  \bibinfo{person}{Liang Lin}.} \bibinfo{year}{2017}\natexlab{}.
\newblock \showarticletitle{Multi-label Image Recognition by Recurrently
  Discovering Attentional Regions}. In \bibinfo{booktitle}{\emph{IEEE
  International Conference on Computer Vision}}. IEEE,
  \bibinfo{pages}{464--472}.
\newblock


\bibitem[\protect\citeauthoryear{Xiao, Xu, Yang, Zhang, Peng, and Zhang}{Xiao
  et~al\mbox{.}}{2015}]%
        {xiao2015application}
\bibfield{author}{\bibinfo{person}{Tianjun Xiao}, \bibinfo{person}{Yichong Xu},
  \bibinfo{person}{Kuiyuan Yang}, \bibinfo{person}{Jiaxing Zhang},
  \bibinfo{person}{Yuxin Peng}, {and} \bibinfo{person}{Zheng Zhang}.}
  \bibinfo{year}{2015}\natexlab{}.
\newblock \showarticletitle{The application of two-level attention models in
  deep convolutional neural network for fine-grained image classification}. In
  \bibinfo{booktitle}{\emph{Proceedings of the IEEE Conference on Computer
  Vision and Pattern Recognition}}. \bibinfo{pages}{842--850}.
\newblock


\bibitem[\protect\citeauthoryear{Xie, Yang, Wang, and Lin}{Xie
  et~al\mbox{.}}{2015}]%
        {xie2014hyper}
\bibfield{author}{\bibinfo{person}{Saining Xie}, \bibinfo{person}{Tianbao
  Yang}, \bibinfo{person}{Xiaoyu Wang}, {and} \bibinfo{person}{Yuanqing Lin}.}
  \bibinfo{year}{2015}\natexlab{}.
\newblock \showarticletitle{Hyper-class augmented and regularized deep learning
  for fine-grained image classification}.
\newblock \bibinfo{journal}{\emph{Proceedings of the IEEE Conference on
  Computer Vision and Pattern Recognition}} (\bibinfo{year}{2015}).
\newblock


\bibitem[\protect\citeauthoryear{Yang, Luo, Change~Loy, and Tang}{Yang
  et~al\mbox{.}}{2015}]%
        {yang2015large}
\bibfield{author}{\bibinfo{person}{Linjie Yang}, \bibinfo{person}{Ping Luo},
  \bibinfo{person}{Chen Change~Loy}, {and} \bibinfo{person}{Xiaoou Tang}.}
  \bibinfo{year}{2015}\natexlab{}.
\newblock \showarticletitle{A large-scale car dataset for fine-grained
  categorization and verification}. In \bibinfo{booktitle}{\emph{Proceedings of
  the IEEE Conference on Computer Vision and Pattern Recognition}}.
  \bibinfo{pages}{3973--3981}.
\newblock


\bibitem[\protect\citeauthoryear{Zhang, Xu, Elhoseiny, Huang, Zhang, Elgammal,
  and Metaxas}{Zhang et~al\mbox{.}}{2016b}]%
        {zhang2016spda}
\bibfield{author}{\bibinfo{person}{Han Zhang}, \bibinfo{person}{Tao Xu},
  \bibinfo{person}{Mohamed Elhoseiny}, \bibinfo{person}{Xiaolei Huang},
  \bibinfo{person}{Shaoting Zhang}, \bibinfo{person}{Ahmed Elgammal}, {and}
  \bibinfo{person}{Dimitris Metaxas}.} \bibinfo{year}{2016}\natexlab{b}.
\newblock \showarticletitle{Spda-cnn: Unifying semantic part detection and
  abstraction for fine-grained recognition}. In
  \bibinfo{booktitle}{\emph{Proceedings of the IEEE Conference on Computer
  Vision and Pattern Recognition}}. \bibinfo{pages}{1143--1152}.
\newblock


\bibitem[\protect\citeauthoryear{Zhang, Donahue, Girshick, and Darrell}{Zhang
  et~al\mbox{.}}{2014}]%
        {zhang2014part}
\bibfield{author}{\bibinfo{person}{Ning Zhang}, \bibinfo{person}{Jeff Donahue},
  \bibinfo{person}{Ross Girshick}, {and} \bibinfo{person}{Trevor Darrell}.}
  \bibinfo{year}{2014}\natexlab{}.
\newblock \showarticletitle{Part-based R-CNNs for fine-grained category
  detection}. In \bibinfo{booktitle}{\emph{European conference on computer
  vision}}. Springer, \bibinfo{pages}{834--849}.
\newblock


\bibitem[\protect\citeauthoryear{Zhang, Xiong, Zhou, Lin, and Tian}{Zhang
  et~al\mbox{.}}{2016a}]%
        {zhang2016picking}
\bibfield{author}{\bibinfo{person}{Xiaopeng Zhang}, \bibinfo{person}{Hongkai
  Xiong}, \bibinfo{person}{Wengang Zhou}, \bibinfo{person}{Weiyao Lin}, {and}
  \bibinfo{person}{Qi Tian}.} \bibinfo{year}{2016}\natexlab{a}.
\newblock \showarticletitle{Picking deep filter responses for fine-grained
  image recognition}. In \bibinfo{booktitle}{\emph{Proceedings of the IEEE
  Conference on Computer Vision and Pattern Recognition}}.
  \bibinfo{pages}{1134--1142}.
\newblock


\bibitem[\protect\citeauthoryear{Zhao, Wu, Feng, Peng, and Yan}{Zhao
  et~al\mbox{.}}{2017}]%
        {zhao2017diversified}
\bibfield{author}{\bibinfo{person}{Bo Zhao}, \bibinfo{person}{Xiao Wu},
  \bibinfo{person}{Jiashi Feng}, \bibinfo{person}{Qiang Peng}, {and}
  \bibinfo{person}{Shuicheng Yan}.} \bibinfo{year}{2017}\natexlab{}.
\newblock \showarticletitle{Diversified Visual Attention Networks for
  Fine-Grained Object Classification}.
\newblock \bibinfo{journal}{\emph{IEEE Transactions on Multimedia}}
  \bibinfo{volume}{19}, \bibinfo{number}{6} (\bibinfo{year}{2017}),
  \bibinfo{pages}{1245--1256}.
\newblock


\bibitem[\protect\citeauthoryear{Zheng, Fu, Mei, and Luo}{Zheng
  et~al\mbox{.}}{2017}]%
        {zheng2017learning}
\bibfield{author}{\bibinfo{person}{Heliang Zheng}, \bibinfo{person}{Jianlong
  Fu}, \bibinfo{person}{Tao Mei}, {and} \bibinfo{person}{Jiebo Luo}.}
  \bibinfo{year}{2017}\natexlab{}.
\newblock \showarticletitle{Learning multi-attention convolutional neural
  network for fine-grained image recognition}. In
  \bibinfo{booktitle}{\emph{Proceedings of the IEEE Conference on Computer
  Vision and Pattern Recognition}}. \bibinfo{pages}{5209--5217}.
\newblock


\end{thebibliography}

\end{document}